\documentclass{article}

\usepackage[final]{neurips_2019}

\usepackage[utf8]{inputenc} 
\usepackage[english]{babel}
\usepackage[T1]{fontenc}    
\usepackage{hyperref}       
\usepackage{url}            
\usepackage{booktabs}       
\usepackage{amsfonts}       
\usepackage{nicefrac}       
\usepackage{microtype}      

\usepackage{amsthm}
\usepackage{amsmath,units}
\usepackage{pgfplots}
\usepackage{amsmath} 
\usepackage{tikz}
\pgfplotsset{compat=1.6}
\usepackage{mathrsfs}
\usepackage[hang, small,labelfont=bf,up,textfont=it,up]{caption} 
\usepackage{booktabs} 
\usepackage{float} 
\usepackage{hyperref} 

\usepackage{etoolbox}
\patchcmd{\thebibliography}{\section*{\refname}}{}{}{}

\usepackage{lettrine} 
\usepackage{paralist} 

\usepackage{titlesec} 
\usepackage{fancyhdr} 
\usepackage{graphicx}
\restylefloat{figure}
\usepackage{url}
\usepackage{amssymb}

\theoremstyle{definition}

\theoremstyle{remark}

\newcommand{\boxedgraphic}[2]{$\vcenter{\hbox{\boxed{\includegraphics[width=#1]{#2}}}}$}

\usepackage{algorithm}
\usepackage[noend]{algpseudocode}

\title{Augmenting Supervised Learning by Meta-learning Unsupervised Local Rules}

\author{%
  Jeffrey Cheng \thanks{Please find my most recent contact information at \texttt{http://jeffreyscheng.com/}.} \\
   Department of Computer Science\\
  University of Pennsylvania\\
  Philadelphia, PA 19104\\
  \texttt{jeffch@seas.upenn.edu}\\
  \And
  Ari Benjamin\\
  Department of Bioengineering\\
  University of Pennsylvania\\
  Philadelphia, PA 19104\\
  \texttt{aarrii@seas.upenn.edu}\\
   \AND
   Benjamin Lansdell\\
   Department of Bioengineering\\
   University of Pennsylvania\\
   Philadelphia, PA 19104\\
   \texttt{lansdell@seas.upenn.edu}
   \And
   Konrad Paul Kording\\
   Department of Bioengineering\\
   University of Pennsylvania\\
   Philadelphia, PA 19104\\
   \texttt{kording@upenn.edu}
}

\begin{document}
\maketitle

\begin{abstract}
    The brain performs unsupervised learning and (perhaps) simultaneous supervised learning. This raises the question as to whether a hybrid of supervised and unsupervised methods will produce better learning. Inspired by the rich space of Hebbian learning rules, we set out to directly learn the unsupervised learning rule on local information that best augments a supervised signal. We present the Hebbian-augmented training algorithm (\textbf{HAT}) for combining gradient-based learning with an unsupervised rule on pre-synpatic activity, post-synaptic activities, and current weights. We test \textbf{HAT}'s effect on a simple problem (Fashion-MNIST) and find consistently higher performance than supervised learning alone. This finding provides empirical evidence that unsupervised learning on synaptic activities provides a strong signal that can be used to augment gradient-based methods.
    
    We further find that the meta-learned update rule is a time-varying function; thus, it is difficult to pinpoint an interpretable Hebbian update rule that aids in training.  We do find that the meta-learner eventually degenerates into a non-Hebbian rule that preserves important weights so as not to disturb the learner's convergence.
\end{abstract}

\section{Prior Work and the Local Meta-Learning Setting}

Backpropagation achieves great performance in neural net optimization, but might not be biologically plausible because most problems are not explicitly phrased as classification with true labels, because neurons only know local signals (e.g. synaptic density, ACh levels, current), and because backpropagation uses the computational graph, a separate data structure with no known biological basis.  

Although some supervised training schemes are more biologically plausible (e.g. contrastive Hebbian learning\cite{seung} and equilibrium propagation\cite{scellier}), it's currently unknown whether the behavior of all neurons is accurately encapsulated by these models.  We speculate that some local, unsupervised learning occurs in the brain and demonstrate that the addition of local, unsupervised rules to standard backpropagation actually improves the speed and robustness of learning. 

\subsection{Local Learning Rules}

We begin by defining a local learning rule.  Consider two adjacent neurons $i, j$ with weight $w_{ij}$: given an impulse traversing $i,j$ with activations $v_i, v_j$, a local learning rule computes updates $\Delta w_{ij}$ using local data $v_i, w_{ij}, v_j$.  Note that by this definition, a local learning rule is unsupervised at face value.

Many neuroscientists have hypothesized specific functions that describe the brain's true (unsupervised) local learning rule.  Most such rules involve using the correlation of activations as part of the update rule.  Examples include Hebb's Rule ~\cite{hebb}, Oja's Rule  ~\cite{oja-1982}, the Generalized Hebbian Algorithm ~\cite{gha}, and nonlineear Hebbian rules ~\cite{nonlinear-hebbian}.

It is not obvious which of these rules (if any) describe the true behavior of neurons.  We employ \emph{meta-learning} (learning how to learn) as an investigative tool.

\subsection{The Meta-Learning Framework}

Optimization functions are algorithms too; it stands to reason that we can learn the best optimization function.  In the meta-learning framework, one model $A$ learns a task (e.g. Fashion-MNIST) while another model $B$ learns how to optimize $A$.  

Meta-learning has achieved great results in finding robust optimization schemes.  Andrychowicz et. al. used meta-learning to find the best gradient-based optimization function ($B$ learns to update $A$ using $A$'s gradients)  \cite{l2lgbg}, and Chen et. al. used meta-learning to find the best gradient-free optimization function ($B$ learns to update $A$ using only the sequence of $A$'s losses).  \cite{l2lwgbg}  Finally, Metz et al. demonstrated a fully differentiable architecture for learning to learn unsupervised local rules and demonstrate better-than-random performance on a few-shot basis. \cite{metz}

If $B$ consistently converges to some stable rule, we take it as strong evidence that this rule may occur in biological brains as well.  We therefore wish to extend Metz's approach to learning semi-supervised local rules not only to improve performance but also to investigate the functional form of the meta-learned update rule.

\section{The Hebbian-Augmented Training Algorithm}

The \textbf{Hebbian-Augmented Training} algorithm (\textbf{HAT}) is an algorithm that trains the neural net $L$ twice per sample: using local, unsupervised rules on the forward pass and using backpropagation-based gradient descent on the backward pass.  

Formally, we create 2 multilayer perceptrons: a learner $L(\cdot\mid\phi_L)$ with parameters $\phi_L$ and a meta-learner $M(v_i, v_j, w_{ij}\mid \phi_M)$ with parameters $\phi_M$, which takes inputs $v_i, w_{ij}, v_j$ and returns $\Delta w_{ij}$.  For a single sample $(\vec{x}, \vec{y})$, we train $L$ without supervision using $M$ and $\vec{x}$; we \textit{simultaneously} train $L$ and $M$ with supervision using $A$ and $\vec{y}$.

\subsection{Phase 1: The Forward Pass}

On the forward pass, we compute activations for each layer.  For a given layer $\ell$, we now have the inputs, outputs, and current weights -- all of the inputs of local learning rule.  We can then apply the outputs of meta-learner $M$ to update the weights of layer $\ell$.  We then \textit{recompute} the activations of layer $\ell$ using the new weights.  This process is done efficiently by convolution (for details, see Appendix A).  We compute the activations of the first layer $\ell_1$, update $\ell_1$, compute the activations of the second layer $\ell_2$, update $\ell_2$, and so on until we compute the predicted Weights $\hat{\vec{y}}$ and update $\ell_{|L|}$.

\subsection{Phase 2: The Backward Pass}

On the backward pass, we backpropagate.  Since we \textit{recomputed} the activations of each layer using weights updated by $M$, the weights of $M$ are upstream of the weights of $L$ in the computational graph; thus, a single iteration of the backpropagation algorithm will compute gradients for both $M$ and $L$.  Given a gradient $\nabla_p$ for each parameter $p\in \phi_L\cup \phi_M$, we then perform a supervised update $p\gets p + A(p, \nabla_p)$.  The key insight is that the convolution of the meta-learner over the weights of the learner forms a fully differentiable framework $M\leadsto L \leadsto \vec{y}$.

\begin{algorithm}[H]
\caption{Hebbian-Augmented Training Algorithm}\label{alg:R}
\begin{algorithmic}[1]
\Procedure{Train-Example}{$L, M, A, \vec{x}, \vec{y}$}
    \State $\vec{v}_0\gets \vec{x}$\Comment{Let $\vec{v}_i$ represent the impulse in layer $i$}
    \For{weights $W_\ell, \ell\in [1...|L|]$} \Comment{Forward pass}
        \State $\hat{v}_{\ell+1}=\sigma(W_\ell\times \vec{v}_{\ell} + b_\ell)$ \Comment{$\hat{v}_{\ell+1}$ is a placeholder output as input to $M$}
        \State $W_\ell\gets W_\ell + M(\vec{v}_\ell, W_\ell, \vec{v}_{\ell+1})$\Comment{Updates weight using local rule $M$}
        \State $\vec{v}_{\ell+1}\gets \sigma(W_\ell\times \vec{v}_{\ell} + b_\ell)$\Comment{Propagate $\hat{v}_{\ell+1}$ as actual layer output}
    \EndFor
    \State Backpropagate loss $H(\vec{v}_{|L|}, \vec{y})$.
    \For{layer weight $W_\ell$ in $L$ and $M$}\Comment{Backward pass}
        \State $W_\ell \gets A\left(\frac{\partial H}{\partial W_\ell}\right)$\Comment{Apply gradient update using optimizer $A$}
    \EndFor
    \State\Return $L, M$\Comment{Return updated learner and updated meta-learner}
\EndProcedure
\end{algorithmic}
\end{algorithm}

\section{\textbf{HAT} Improves Performance on Fashion-MNIST}

We hypothesize that the \textbf{HAT} algorithm will have three positive effects.
\begin{itemize}
    \item \textbf{HAT} will train the learner $L$ faster since there are twice as many updates.  In ordinary backpropagation the metadata generated from the forward pass is computed and wasted; in \textbf{HAT}, the metadata is computed and used to generate a (potentially) useful update.
    \item \textbf{HAT} will improve the convergence of $L$.  The second update should introduce some stochasticity in the loss landscape since it is not directly tied to gradient descent, which may lead $L$ into better local optima.
    \item \textbf{HAT} will improve the performance of $L$ when some examples are not labeled.  Backpropagation has no ability to learn from just the input $\vec{x}$, while \textbf{HAT} is able to perform the unsupervised update.
\end{itemize}

We generate two learning curves to test these hypotheses: one with respect to time and one with respect to the proportion of labeled examples.  The charts below represent the aggregated learning curves of 100 pairs $(L_i, M_i)$.

\begin{figure}[H]
    \centering
    \boxedgraphic{0.45\textwidth}{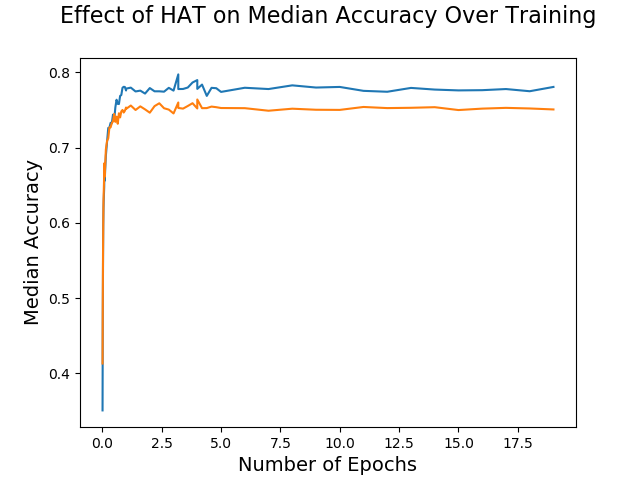}
    \boxedgraphic{0.45\textwidth}{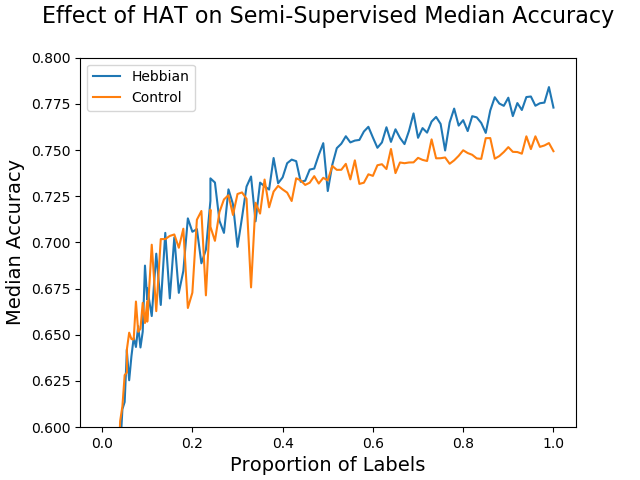}
    \caption{The effect of \textbf{HAT}'s on median accuracy curves.}
\end{figure}

\vspace*{0.2cm}

We find that the effects of \textbf{HAT} on training are clearly positive. The median accuracy of the neural nets trained by \textbf{HAT} is clearly increased along the learning curve, and the \textbf{HAT}-group neural nets reach a higher asymptotic value than the control group.  We do note that the two learning curves seem to inflect around the same point -- \textbf{HAT} does not seem to cause a faster convergence, just a better one.  We attribute this to the meta-learner's convergence; it may take the meta-learner up to 0.5 epochs to start to have positive effects.

One potential concern with adding unsupervised meta-learner updates is that after the convergence of the base learner $L$, the meta-learner's continued output of non-zero updates might ``bounce'' the base learner out of an optimum.  Remarkably, we see in the above plot that the performance of the \textbf{HAT}-trained neural nets is quite stable for the entire 18 epochs of post-convergence duration.

To our surprise, we find that \textbf{HAT} is more effective when there are more labels, even though the self-supervised component of the algorithm is designed to take advantage of scarce labels.  We attribute this to slow convergence of the meta-learner $M$ -- when labels are scarce, the meta-learner may actually converge slower than the learner and thus provide bad update suggestions.

\section{The Behavior of the Meta-Learned Update Rule}

We would like insight into why \textbf{HAT} improves the training of neural nets over vanilla gradient descent.  Thus, we will analyze the functional form of the learned update rule $M$ after it has fully converged.  Recall the setting from experiments 1 and 2: we generate 100 pairs of learners and meta-learners: $(L_i, M_i)$ for $i\in \{1,...,100\}$.  We then investigate the pointwise mean $\overline{M} $of these meta-learners. 

We first visualize the dependence of the function $\overline{M}$ on its inputs $(v_i, v_j, w_{ij})$. 
\begin{figure}[H]
    \centering
    \boxedgraphic{0.45\textwidth}{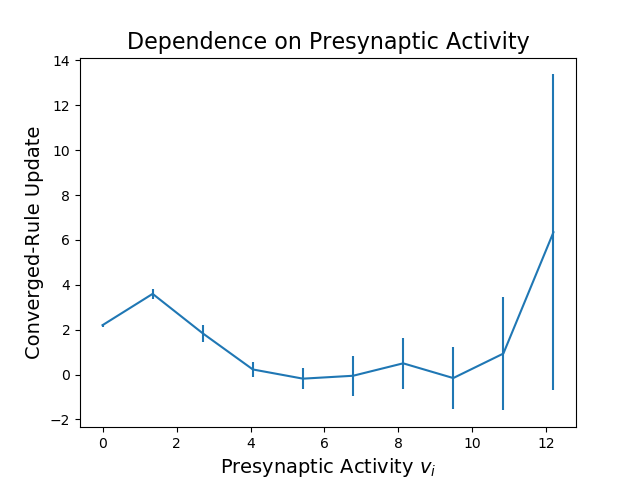}
    \boxedgraphic{0.45\textwidth}{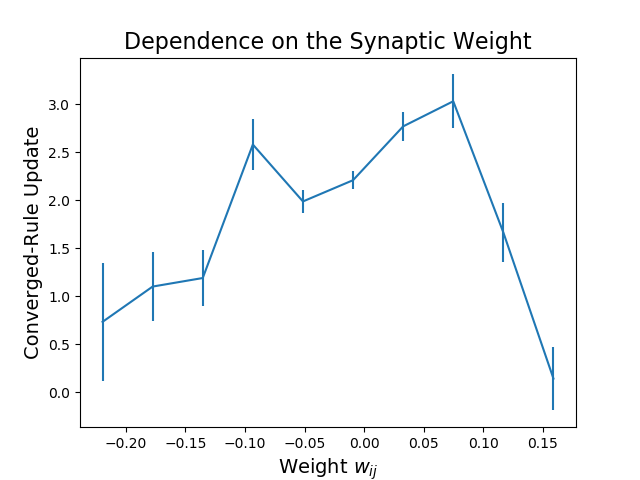}
    \boxedgraphic{0.45\textwidth}{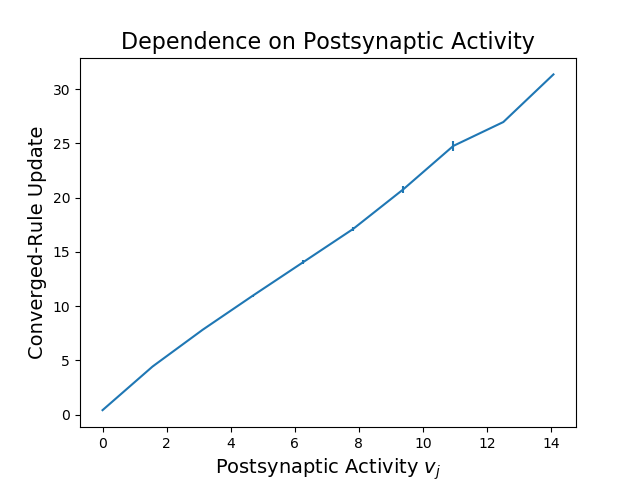}
    \caption{These plots show very little dependence of the converged rule on $v_i$ and $w_{ij}$.}
    \label{fig:my_label}
\end{figure}

We find that a remarkably linear dependence on $v_j$ explains almost all of the variance in the outputs of the meta-learned update rule.  This indicates that the rule is a ``rich-get-richer'' scheme: neurons that already fired with high magnitude will experience larger incoming weights and thus be encouraged to fire with high activation in the future.

This linear dependence is surprising since all of the hypothesized rules in neuroscience have a dependence on $v_i\cdot v_j$.  As a sanity check, we attempted to directly apply this update rule ($\Delta w_{ij} \approx 2\cdot v_j$) without meta-learning to see if we can replicate \textbf{HAT}'s performance improvement.  However, the results were decisively negative -- \textbf{HAT} improves performance, but the a priori application of \textbf{HAT}'s update rule decreases it.  We present three hypotheses:

\begin{itemize}
    \item Perhaps $M$ learns a good update rule while $L$ is training, then learns a degenerate rule once $L$ has converged.  The sole purpose of this degenerate rule would be to not un-learn the important weights that have already converged (thus explaining the rich-gets-richer behavior of the rule $f(\cdot)=2v_j$).  Thus, analyzing the black-box function at epoch 20 is merely the wrong time -- perhaps observing the meta-learned rule at epoch 1 would be more insightful and useful.
    \item Perhaps $M$ learns a good update rule in each run, and these update rules are all complex functions with no good low-order polynomial approximations; however, their pointwise mean (which is itself not a good local update rule) happens to be linear.  Thus, $\overline{M}$ is the wrong object to analyze and presents behaviors that are not indicative of the results of experiments 1 and 2.
    \item Perhaps the learning of $M$ is extremely transient.  For any given point in time, there is a different optimal learning rule, and our exercise in finding a fixed local, unsupervised update rule that is universal across training is futile.
\end{itemize}

\section{Conclusion}

The \textbf{HAT} algorithm demonstrates that local, unsupervised signals can provide performance-improving weight updates.  Neural nets under \textbf{HAT} converge to better asymptotic losses as long as there is sufficient time ($>0.5$ epochs) and a sufficient number of labels ($>20\%$ of the data is labeled).  The latter finding is surprising since the addition of an unsupervised learning algorithm depends on the presence of labels in order to deliver marginal benefits over gradient descent.

The underlying form of the learned rule that makes \textbf{HAT} successful is still a mystery; we find that while the meta-learner may learn a useful update rule during training, the meta-learner does not converge to this useful rule in the long run and instead devolves into a linear function \textbf{Converged-Rule}.  This converged function preserves fully-converged weights by reinforcing incoming weights for neurons with high activations.

\subsection{Future Work}

The discovery that \textbf{HAT} does not stably converge to a function makes analysis quite difficult.  However, there is potential for future work to do more subtle analyses.

Imagine a time $t$ during training in which the meta-learner $M$ has converged to a useful function, but the learner $L$ has not yet finished training.  A follow-up to this thesis might be to discover whether there such a time $t$ exists, what the structure of $M$ at time $t$ is, and how $M$ changes the weights of $L$ at time $t$.  One potential methodology might be to observe the function $f$ not as a 3-dimensional function in $(v_i, w_{ij}, v_j)$ but rather as a 4-dimensional function in $(v_i, w_{ij}, v_j, t)$.  Observing the function along the $t$-axis and checking for phase changes would shed light on whether a single useful update rule is learned during training or whether \textbf{HAT}'s learning is truly transient and continuous.  If this follow-up were to succeed, then we could have an a priori rule to apply without having to metalearn update rules.

Extracting the local rules from multiple domains could either find that \textbf{HAT} learns a universal rule or that functional distance between two rules describes the ``difference'' between their originating domains.
\vspace*{-1mm}
\begin{itemize}
    \itemsep-0.4em 
    \item Suppose we always metalearn the same rule, regardless of problem domain.  \textbf{Optimal-Hebb} is then a universal learning rule.
    \item Suppose \textbf{Optimal-Hebb} is not universal for all problems.  For local rules $R_A, R_B$ on problems $A,B$, integrating $\int_{\mathbb{R}^3} (R_A-R_B)\cdot dF(v_i, w_{ij}, v_j)$ for input distribution $F$ gives an explicit measure for how similar $A$ and $B$ are.  This provides a systematic way to identify pairs of learning problems that are good candidates for transfer learning.
\end{itemize}

\newpage
\bibliographystyle{plain}
\bibliography{refs.bib}

\newpage
\section{Appendices}

\subsection{Appendix A}

One implementation detail is notably not covered in the \textbf{HAT} pseudocode; this implementation detail patches an inadequacy in modern deep learning frameworks.

Given two neural net layers $\ell_i$ and $\ell_{i+1}$ and minibatches of size $B$, we have $B$ instances of $|\ell_i|\times \ell_{i+1}$ neuron pairs, each of which has 3 salient properties ($v_i, w_{ij}, v_j$).  Therefore, we would like to apply the function $M$ over the zeroth dimension of a tensor of size $3\times B\times |\ell_i|\times |\ell_{i+1}|$ in order to compute the unsupervised weight updates.  

However, as of this writing date, it is not possible to apply an arbitrary function $M$ to slices of a tensor in parallel in \textbf{any} modern deep learning framework (e.g. Tensorflow, PyTorch, Keras); the reason is that this plays poorly with optimization of the computational graph.  We thus implement the application of $M$'s updates to the weights by convoluting $M$ over a state tensor.

This is best clarified with an example.  Suppose we have a neural net with consecutive layers $\ell_1, \ell_2$ of size 784 and 183, respectively.  Suppose further that we have batches of size 50.  Finally, suppose that we require a meta-learner that is a neural net of architecture $3\times 100\times 1$.

For a single batch, we have the tensors of the following sizes:
\begin{align*}
    \vec{v_i}: 50\times 784 ~~~&\rightarrow~~~ 50\times \boxed{1}\times 784\\
    \vec{w_{ij}}: 183\times 784~~~&\rightarrow~~~ \boxed{1} \times 183\times 784 \\
    \vec{v_j}: 50\times 183 ~~~&\rightarrow~~~ 50\times 183\times \boxed{1}
\end{align*}
We then copy the tensors along the boxed dimensions to stack them.
\begin{align*}
    \vec{v_i}: 50\times \boxed{1}\times 784 ~~~\rightarrow~~~ &50\times \boxed{183}\times 784\\
    \vec{w_{ij}}: \boxed{1} \times 183\times 784~~~\rightarrow~~~ &\boxed{50} \times 183\times 784 \\
    \vec{v_j}: 50\times 183\times \boxed{1}~~~\rightarrow~~~ &50\times 183\times \boxed{784}\\
    \hline\\
    \text{Input to to meta-learner:} ~~~&\boxed{3\times 50\times 183\times 784}
\end{align*}
We instantiate $M$ as a sequence of 3 composed functions:
\begin{enumerate}
    \item a convolutional layer of kernel size $1\times 1$ with 3 in-channels and 100 out-channels,
    \item a ReLU activation, and
    \item a convolutional layer of kernel size $1\times 1$ with 100 in-channels and 1 out-channels.
\end{enumerate}
Applying this series of functions to a $1\times $ image with 3 channels is equivalent to passing the 3 channels into a neural net with architecture $3\times 100\times 1$.

PyTorch (the framework used for this research) does not support the vectorization of arbitrary functions along torch tensors.  However, it \textbf{does} support (and heavily optimize for) convolutions. Thus, we implement our neural net function $M$ as a series of convolutions, and we convolve the function over the input tensor of size $3\times 50\times 183\times 784$.  The output of $M$ is of size $50\times 183\times 784$; we average over the zeroth dimension to finally get a weight update of dimension $183\times 784$, which is the same size as the original weight tensor.

\end{document}